\pdfoutput=1

\documentclass[11pt]{article}

\usepackage{emnlp2021}

\usepackage{times}
\usepackage{latexsym}

\usepackage{times}
\usepackage{latexsym}
\usepackage{graphicx}

\usepackage{amsmath}
\usepackage{cleveref}
\usepackage{xcolor}
\usepackage{enumitem}
\usepackage{tikz}
\usepackage{pgfplots}
\pgfplotsset{compat=newest}
\usepackage{subcaption}
\usepackage{url}
\usepackage{booktabs}
\usepackage{microtype}
\usepackage{placeins}

\usepackage[T1]{fontenc}

\usepackage[utf8]{inputenc}

\usepackage{microtype}

\title{How May I Help You? \\
Using Neural Text Simplification to Improve Downstream NLP Tasks}

\author{Hoang Van, Zheng Tang, \and Mihai Surdeanu \\
  Department of Computer Science, University of Arizona \\
  \texttt{\{vnhh,zhengtang,msurdeanu\}@email.arizona.edu}}

\begin{document}
\maketitle
\begin{abstract}
The general goal of text simplification (TS) is to reduce text complexity for human consumption. In this paper, we investigate another potential use of neural TS: assisting machines performing natural language processing (NLP) tasks. We evaluate the use of neural TS in two ways: simplifying input texts at prediction time and augmenting data to provide machines with additional information during training. We demonstrate that the latter scenario provides positive effects on machine performance on two separate datasets. In particular, the latter use of TS significantly improves the performances of LSTM (1.82--1.98\%) and SpanBERT (0.7--1.3\%) extractors on TACRED, a complex, large-scale, real-world relation extraction task. Further, the same setting yields significant improvements of up to $0.65\%$ matched and $0.62\%$ mismatched accuracies for a BERT text classifier on MNLI, a practical natural language inference dataset.
\end{abstract}

\section{Introduction}
The goal of text simplification (TS) is to reduce text complexity (while preserving meaning) such that the corresponding text becomes more accessible to human readers. Previous works explored how TS can assist children \cite{kajiwara2013selecting}, non-native speakers \cite{pellow2014open}, and people with disabilities \cite{rello2013impact}. While this can be achieved in a variety of approaches \cite{sikka2020survey}, most TS research has focused on two major approaches: rule-based and neural sequence-to-sequence (seq2seq). Since 2017, there is a significant increase of neural seq2seq TS methods \cite{zhang2017sentence,zhao2018integrating,kriz2019complexity,maddela2020controllable,jiang2020neural}. 

In this paper, we analyze another potential use of the latter TS direction: assisting machines performing natural language processing (NLP) tasks. To this end, we investigate two possible directions: (a) using TS to simplify input texts at prediction time, and (b) using TS to augment training data for the respective NLP tasks. We empirically analyze these two directions using two neural TS systems \cite{martin2019controllable,nisioi2017exploring}, and two NLP tasks: relation extraction using the TACRED dataset \cite{zhang2017position}, and multi-genre natural language inference (MNLI) \cite{williams2017broad}. Further, within these two tasks, we explore three methods: two relation extraction approaches, one based on LSTMs \cite{hochreiter1997long} and another based on transformer networks, SpanBERT \cite{joshi2020spanbert}, and one method for MNLI also based on transformer networks, BERT \cite{devlin2018bert}.


Our analysis shows that simplifying texts at prediction times does not improve results, but using TS to augment training data consistently helps in all configurations. 
In particular, after augmented data is added, all approaches outperform their respective configurations without augmented data on both TACRED (0.7--1.98\% in F1) and MNLI (0.50--0.65\% in accuracies) tasks. The reproducibility checklist and the software are available at this link: \url{https://github.com/vanh17/TextSiM}.

\section{Related Work}
Recent work have effectively proven the practical application of neural networks and neural deep learning approaches to solving machine learning problems \cite{ghosh2021ensemble,blalock2020state,yin2017comparative}. 

 With respect to input simplification, several works have utilized TS as a pre-processing step for downstream NLP tasks such as information extraction \cite{miwa2010entity,schmidek2014improving,niklaus2017sentence}, parsing \cite{chandrasekar-etal-1996-motivations}, semantic role labeling \cite{vickrey2008sentence}, and machine translation \cite{vstajner2016can}. However, most of them focus on the use of rule-based TS methods. In contrast, we investigate the potential use of domain-agnostic neural TS systems in simplifying inputs for downstream tasks. 
We show that, despite the complexity of the tasks investigated and the domain agnosticity of the TS approaches, TS improves both tasks when used for training data augmentation, but not when used to simplify evaluation texts.

On data augmentation for natural language processing downstream tasks, previous work show significant benefits of introducing noisy data on the machine performance \cite{van2021cheap,kobayashi2018contextual}. Previous efforts used TS approaches, e.g. lexical substitution, to augment training data for downstream tasks such as text classification \cite{zhang2015character,wei2019eda}. However, these methods focused on replacing words with thesaurus-based synonyms, and did not emphasize other important lexical and syntactic simplification. Here, we use two out-of-the-box neural TS systems that apply both lexical and syntactic sentence simplification for data augmentation, and show that our data augmentation consistently leads to better performances. Note that we do not use rule-based TS systems because they have been proven to perform worse than their neural counterparts \cite{zhang2017sentence,nisioi2017exploring}. Further, rule-based TS systems are harder to build in a domain-independent way due to the many linguistic/syntactic variations across domains.

\section{Approach}
We investigate the impact of text simplification on downstream NLP tasks in two ways: (a)
simplifying input texts at prediction time, and (b) augmenting training data for the respective NLP tasks. We discuss the settings of these experiments next. 

\begin{table}[t]
\centering
\scalebox{.9}{%
\begin{tabular}{l l c c}
\toprule
& & ACCESS & NTS \\
\midrule
1 & Training Data  &     0.67 $\pm$ 0.16 & 0.89 $\pm$ 0.22  \\
\midrule
2 & Dev Data    & 0.68 $\pm$ 0.15  & 0.92 $\pm$ 0.18  \\
\bottomrule
\end{tabular}}
\caption{The empirical differences in BLEU scores \cite{papineni2002bleu} between original and simplified text generated by two TS systems, ACCESS and NTS, in TACRED training and dev datasets.} 
\label{tab:tacred_bleu}
\end{table}

\begin{table}[t]
\centering
\scalebox{0.85}{%
\begin{tabular}{l l c c}
\toprule
& & ACCESS & NTS \\
\midrule
& \bf{\emph{Train}}  &         &       \\
\midrule
1 & Premise    & 0.62 $\pm$ 0.24  &  0.76 $\pm$ 0.25 \\ 
\midrule
2 & Hypothesis & 0.62 $\pm$ 0.30 & 0.80 $\pm$ 0.17 \\
\midrule
& \bf{\emph{Dev mismatched}} &  &  \\
\midrule
3 & Premise    &    0.62 $\pm$ 0.28 & 0.80 $\pm$ 0.22 \\
\midrule
4 & Hypothesis & 0.65 $\pm$ 0.23 & 0.81 $\pm$ 0.17 \\
\midrule
& \bf{\emph{Dev matched}} &  &  \\
\midrule
5 & Premise    &  0.62 $\pm$ 0.30 & 0.75 $\pm$ 0.26 \\
\midrule
6 & Hypothesis & 0.60 $\pm$ 0.25 & 0.80 $\pm$ 0.17 \\
\bottomrule
\end{tabular}}
\caption{The empirical differences in BLEU scores \cite{papineni2002bleu} between original and simplified text generated by two TS systems, ACCESS and NTS, in MNLI training and dev datasets.} 
\label{tab:mnli_bleu}
\end{table}

\begin{table}[t]
\centering
\scalebox{0.80}{%
\begin{tabular}{l l c c c}
\toprule
& & Simplified & Original & Simplified + \\
& &          &           & Complement \\
\midrule
& \bf{\emph{Train Data Sets}}  &         &       \\
\midrule
& \bf{LSTM}    &         &       \\  
\midrule
1 & Original & 59.95 & 62.70 & 61.32\\
\midrule
2 & Simplified +  & 62.34 & 62.59 & 62.12\\
 &  Complement & & & \\
\midrule
3 & Simplified +  & \bf{62.64} & \bf{64.52} & \bf{64.08} \\
  &  Original (AD)          &       &       & \\
\midrule
& \bf{SpanBERT}    &         &       \\  
\midrule
4 & Original & 62.42 & 66.70 & 64.12 \\
\midrule
5 & Simplified +  & 64.12 & 65.45 & 64.92 \\
 &  Complement & & & \\
\midrule
6 & Simplified +  & \bf{65.14} & \bf{68.00} & \bf{66.43} \\
  &  Original (AD)          &       &       & \\
\bottomrule
\end{tabular}}
\caption{F1 on TACRED test set of the LSTM and SpanBERT approaches using ACCESS \cite{martin2019controllable} as the TS method. The different rows indicate the different data augmentation strategies applied on the training data, while the columns indicate the type of simplification applied at runtime on the test data. We investigated the following configurations:
{\em Original}: unmodified dataset; 
{\em Simplified + Complement}: consists of simplified data that preserves critical information combined with original data when simplification fails to preserve important information; {\em Simplified + Original}: consists of all original data augmented with additional simplified data that preserves critical information. {\em (AD)} annotates models using data augmented by neural TS systems during training.} 
\label{tab:results_tacred_access}
\end{table}

\subsection{Input Simplification at Prediction Time \label{sec:input_simplification}}
We pose the run-time input simplification problem as a transparent data pre-processing problem.
That is, given an input data point, we simplify the text while keeping the native format of the task, and then feed the modified input to the actual NLP task.
For example, for the TACRED sentence \emph{``the CFO Douglas Flint will become chairman, succeeding Stephen Green is leaving for a government job.''}, which contains a per:title relation between the two entities {\em Douglas Flint} and {\em chairman},  our approach will first simplify the text to \emph{``the CFO Douglas Flint will become chairman, and Stephen Green is leaving to take a government job.''}. Then we generate a relation prediction for the simplified text using existing relation extraction classifiers.

\subsection{Data Augmentation for Training \label{sec:data_augmentation}}
Here, we augment training data by simplifying the text of some original training examples, and appending it to the original training dataset. First, we sample which examples should be used for augmentation with probability $p$. Second, once an example is selected for augmentation, we generate an additional example with the text portion simplified using TS. For example, for the data in section \ref{sec:input_simplification}, we generate an additional training data with the corresponding simplified text. $p$ is a hyper parameter that we tuned for each task (see next section).
\begin{table}[t]
\centering
\scalebox{0.80}{%
\begin{tabular}{l l c c c}
\toprule
& & Simplified & Original & Simplified + \\
& &          &           & Complement \\
\midrule
& \bf{\emph{Train Data Sets}}  &         &       \\
\midrule
& \bf{LSTM}    &         &       \\  
\midrule
1 & Original & 60.47 & 62.70 & 61.03\\
\midrule
2 & Simplified +  & \bf{63.40} & 62.96 & 62.28 \\
 &  Complement & & & \\
\midrule
3 & Simplified +  & 62.91 & \bf{64.68} & \bf{64.35} \\
  &  Original (AD)          &       &       & \\
\midrule
& \bf{SpanBERT}    &         &       \\  
\midrule
4 & Original & 62.20 & 66.70 & 63.90 \\
\midrule
5 & Simplified +  & 64.12 & 65.32 & 63.92 \\
 &  Complement & & & \\
\midrule
6 & Simplified +  & \bf{65.32} & \bf{67.40} & \bf{65.47} \\
  &  Original (AD)           &       &       & \\
\bottomrule
\end{tabular}}
\caption{ F1 on TACRED test set of the LSTM and SpanBERT approaches using NTS \cite{nisioi2017exploring} as the TS method. This table follows the same format as Table~\ref{tab:results_tacred_access}.} 
\label{tab:results_tacred_nts}
\end{table}

\begin{table}[t]
\centering
\scalebox{0.80}{%
\begin{tabular}{l l c c}
\toprule
& & Simplified & Original \\
& & m/mm acc   & m/mm acc   \\
\midrule
& \bf{\emph{Train Data Sets}}  &         &       \\
\midrule
& \bf{ACCESS}    &   &   \\ 
\midrule
1 & Original & 71.25/71.43 & 82.89/83.10 \\
\midrule
2 & Original Swapped   & 71.76 $\pm$ 0.13/ & 83.00 $\pm$ 0.03/ \\
 & with 10\% Simplified & 72.12 $\pm$ 0.08 & 83.25 $\pm$ 0.05 \\
\midrule
3 & Original Swapped   & 72.74 $\pm$ 0.10/ & 82.66 $\pm$ 0.07/ \\
 & with 20\% Simplified & 73.10 $\pm$ 0.12 & 82.88 $\pm$ 0.09 \\
\midrule
4 & 5\% Simplified & 71.30 $\pm$ 0.15/ & \bf{83.47} $\pm$ \bf{0.04} \\
  & + Original (AD) & 71.52 $\pm$ 0.10 & \bf{83.61} $\pm$ \bf{0.08}\\
\midrule
5 & 10\% Simplified  & 71.81 $\pm$ 0.07/ & 82.81 $\pm$ 0.05/ \\
  & + Original (AD)      & 71.99 $\pm$ 0.08 & 83.05 $\pm$ 0.09 \\
\midrule
6 & 15\% Simplified  & 71.87 $\pm$ 0.11/ & 82.92 $\pm$ 0.05/ \\
  & + Original (AD)  & 72.10 $\pm$ 0.07 & 83.13 $\pm$ 0.06 \\
\midrule
& \bf{NTS}    &     &  \\  
\midrule
7 & Original & 33.36/33.53 & 82.89/83.10 \\
\midrule
8 & Original Swapped   & 33.39 $\pm$ 0.10/ & 83.28 $\pm$ 0.07/ \\
 & with 10\% Simplified & 33.46 $\pm$ 0.08 & 83.50 $\pm$ 0.11 \\
\midrule
9 & Original Swapped   & 33.71 $\pm$ 0.08/ & 82.60 $\pm$ 0.14/ \\
 & with 20\% Simplified & 33.90 $\pm$ 0.11/ & 82.79 $\pm$ 0.09 \\
\midrule
10 & 5\% Simplified & 33.35 $\pm$ 0.10/ & 83.20 $\pm$ 0.09/ \\
  & +  Original (AD) & 33.50 $\pm$ 0.09 & 83.41 $\pm$ 0.10 \\
\midrule
11 & 10\% Simplified & 33.50 $\pm$ 0.07/ & \bf{83.51} $\pm$ \bf{0.05}/ \\
  &  + Original (AD) & 33.80 $\pm$ 0.09 & \bf{83.70} $\pm$ \bf{0.07} \\
\midrule
12 & 15\% Simplified  & 33.65 $\pm$ 0.04/ & 83.09 $\pm$ 0.05 \\
  &  + Original (AD)  & 33.79 $\pm$ 0.10  &  83.25 $\pm$ 0.07\\
\bottomrule
\end{tabular}}
\caption{\footnotesize Matched (m) and mismatched (mm) accuracies on MNLI development 
using text simplified/augmented by ACCESS (top half) and NTS (bottom half).
{\em Original Swapped with x\% Simplified} consists of original data with x\% of data points replaced with their simplified form. {\em x\% Simplified + Original} consists of the original data augmented with an additional x\% of simplified data. {\em (AD)} annotates models using data augmented by neural TS systems during training. Note that our results in the original configuration differ slightly from those in \cite{devlin2018bert}. This is likely due to the different hardware and library versions used \cite{belz2021systematic}.}
\label{tab:results_mnli_dev}
\end{table}

\section{Experimental Setup}
\paragraph{NLP tasks and methods:}
We evaluate the impact of TS on two NLP tasks: (a) relation extraction (RE) using the TACRED dataset \cite{zhang2017position}, and (b) natural language inference (NLI) on the MNLI dataset \cite{williams2017broad}. 

TACRED is a large-scale RE dataset with 106,264 examples built on newswire and web text with an average sentence length of 36.4 words. Each sentence contains two entities in focus (called subject and object) and a relation that holds between them. We selected this task because the nature of RE requires critical information  preservation, which is challenging for neural TS methods \cite{van2020automets}. That is, the simplified sentences must contain the subject and object entities. 

The MNLI corpus is a crowd-sourced collection of 433K sentence pairs annotated for NLI. The average sentence length in this dataset is 22.3 words. Each data point contains a premise-hypothesis pair and one of the three labels: contradiction, entailment, and neutral. We selected MNLI as the second task to further understand the effects of TS on machine performance on tasks that rely on long text, which is a challenge for TS methods \cite{shardlow2014survey,xu2015problems}. 

We train three approaches for these two tasks. First, for TACRED, we use a classifier based on LSTMs\footnote{\url{https://github.com/yuhaozhang/tacred-relation}} \cite{hochreiter1997long}, and a second based on SpanBERT\footnote{\url{https://huggingface.co/SpanBERT/spanbert-large-cased}} \cite{joshi2020spanbert}.
For MNLI, we trained a BERT-based classifier\footnote{\url{https://huggingface.co/bert-base-cased}} \cite{devlin2018bert}. 
For reproducibility, we use the default settings and general hyper parameters recommended by the task and creators of the transformer networks \cite{zhang2017position, joshi2020spanbert,devlin2018bert}. Through this, we aim to separate potential improvements of our approaches from those coming from improved configurations.

\begin{table}[t]
\centering
\scalebox{0.85}{%
\begin{tabular}{l l c c}
\toprule
& & Simplified & Original \\
& & m/mm acc   & m/mm acc   \\
\midrule
& \bf{\emph{Train Data Sets}}  &         &       \\
\midrule
& \bf{ACCESS}    &   &   \\  
\midrule
1 & Original & 71.10/71.30 & 82.78/83.00 \\
\midrule
2 & 5\% Simplified + & \bf{71.21}/\bf{71.40} & \bf{83.37}/\bf{83.50} \\
 & Original (AD) & & \\
\midrule
& \bf{NTS}    &      \\ 
\midrule
3 & Original & 33.25/33.45 & 82.78/83.00 \\
\midrule
4 & 10\% Simplified +  & \bf{33.39}/\bf{33.61} & \bf{83.43}/\bf{83.62} \\
 & Original (AD) & & \\
\bottomrule
\end{tabular}}
\caption{ Matched (m) and mismatched (mm) accuracies on MNLI test, using the best configurations from development.} 
\label{tab:results_mnli_test}
\end{table}

\paragraph{Text simplification methods:}
For TS, we use two out-of-the-box neural seq2seq TS approaches: ACCESS \cite{martin2019controllable},  and NTS \cite{nisioi2017exploring}. Tables~\ref{tab:tacred_bleu} and \ref{tab:mnli_bleu} show the BLEU scores \cite{papineni2002bleu} between original and simplified text generated by these two TS systems for the two tasks. The tables highlight that both systems change the input texts, with ACCESS being more aggressive. 

\begin{table*}
\scalebox{.745}{%
\begin{tabular}{l l l l}
\toprule
 & \textbf{Gold Data} & \textbf{Our Approach} & \textbf{Baseline} \\
\midrule
 1 & \textbf{P:} In the apt description of one witness, it drops below the  & \textbf{P:} It drops below the radar screen ... you don't know & \textbf{Predict:} \\
  & radar screen ... you don't know where it is.  \textbf{H:} It is hard & where it is. \textbf{H:} It is hard for one to find & Neutral \\
    & for one to realize  what just happened. \textbf{Gold Label:} Entailment & what just happened. \textbf{Predict:} Entailment & \\
\midrule
 2 & \textbf{P:} The tourist industry continued to expand,  and though it   & \textbf{P:} The tourist industry continued to expand, and & \textbf{Predict:} \\
 
   & the top two income earners in Spain, was ... consequences. & ... top two income earners in Spain. ... consequences. & Contradict \\
   
   & \textbf{H:} Tourism is not very big in Spain. \textbf{Gold Label:} Contradict & \textbf{H:} Tourism is very big in Spain. \textbf{Predict:} Entailment & \\

\midrule
3 & \textbf{P:} This site includes a list of all award winners and a & \textbf{P:} This site includes a list of all award winners and a & \textbf{Ans:} \\

  & searchable database of Government Executive articles. & searchable database of Government Executive articles.  & Neutral \\
  
  & \textbf{H:} The Government Executive articles housed on the website & \textbf{H:} The Government Executive articles are not able & \\
  
  & are not able to be searched. \textbf{Ans:} Contradict & to find the website to be searched . \textbf{Ans:} Contradict &  \\
  
\bottomrule
\end{tabular}}
\caption{ Qualitative comparison of the outputs from our approach (text simplification by ACCESS) and the respective BERT baseline on the original MNLI data. \emph{P, H} indicates premise and hypothesis.
}
\label{tab:example_mnli}
\end{table*}

\paragraph{Evaluation measures:}
We directly followed the evaluation measures proposed by the original task organizers \cite{zhang2017position, williams2017broad}. Specifically, we used these main  metrics: (a) F1 on TACRED relation extraction, and (b) matched/mismatched accuracies on MNLI.

\paragraph{Hyper parameter tuning:}
We tuned the only hyper parameter for data augmentation, the percentage of augmented data points, $p$, for MNLI. On this task we augmented 5, 10, and 15\% of sentence pairs from training data, and found 5 and 10\% of training data as the best thresholds for ACCESS and NTS respectively. 
For TACRED, we did not use this hyper parameter. Instead, we used all simplifications that preserve critical information for data augmentation. That is, we added all simplified sentences that preserve the subject and object entities necessary for the underlying relation. We found that 66\% of training data sentences could be simplified while preserving this information by ACCESS, and 72\% by NTS.

\section{Results and Discussion}

Tables~\ref{tab:results_tacred_access} and \ref{tab:results_tacred_nts} summarize our results on TACRED for the two distinct TS methods. Because we tuned the hyper parameter $p$ for MNLI, we report results on both development and test for this task (Tables~\ref{tab:results_mnli_dev} and \ref{tab:results_mnli_test}, respectively). Further, for MNLI we also report average performance (and standard deviation) for 3 runs, where we select a different sample to be simplified in each run. This is not necessary for TACRED; for this task we simplified {\em all} data points that preserved critical information i.e., the two entities participating in the relation.\footnote{This is not possible for MNLI, where it is unclear which part of the text is critical for the task.}

\paragraph{Input simplification at prediction time:} Tables~\ref{tab:results_tacred_access} and \ref{tab:results_tacred_nts} show that simplifying inputs at test time does not yield improvements (compare the {\em Original} column with the third one).
There are absolute decreases in performance of 1.38--2.58\% and 1.67--2.80\% in F1 on TACRED for ACCESS and NTS systems, respectively (substract column 3 from column 2 in rows 1 and 4). 

Similarly, on MNLI, the performance on simplified inputs is lower than the classifier tested on the original data. The performance drops on MNLI are more severe (11.68--49.53\% and 11.70--49.55\% in matched and mismatched accuracies) (substract column 1 from column 2 in row 1 and row 3 in Table \ref{tab:results_mnli_test} pairwise). We hypothesize that this is due to the quality of simplifications in MNLI being lower than those in TACRED. In the latter situation we could apply a form of quality control, i.e., by accepting only the simplifications that preserve the subject and object of the relation. 
To illustrate the benefits/dangers of text simplification, we show a few examples where simplification improves/hurts MNLI output in Table~\ref{tab:example_mnli}. 


\paragraph{Augmenting training data:} As shown in row 3 and 6 in Table~\ref{tab:results_tacred_access} and \ref{tab:results_tacred_nts}, all methods trained on augmented data yield consistent performance improvements, regardless of the RE classifier used (LSTM or SpanBERT) or TS method used (ACCESS or NTS). 
There are absolute increases of 1.30--1.82\% F1 for ACCESS and 0.70--1.98\% F1 for NTS on (substract row 1 from row 3 and row 4 from row 6 for ACCESS and NTS respectively). 
The best configuration is when the original training data is augmented with all data points that could be simplified while preserving the subject and object of the relation (rows 4 and 8 in the two tables).
These results confirm that TS systems can provide additional, useful training information for RE methods. 

Similarly, on MNLI, the classifier trained using augmented data outperforms the BERT classifier that is trained only on the original MNLI data. For two TS systems, ACCESS and NTS, we observe performance increases of 0.59--0.65\% matched accuracy, and 0.50--0.62\% mismatched accuracy (compare rows 1 vs. 2, and row 3 vs. 4 in Table \ref{tab:results_mnli_test}). This confirms that TS as data augmentation is also useful for NLI.

All in all, our experiments suggest that our data augmentation approach using TS is fairly general. It does not depend on the actual TS method used, and it improves three different methods from two different NLP tasks. Further, our results indicate that our augmentation approach is more beneficial for tasks with lower resources (e.g., TACRED), but its impact decreases as more training data is available (e.g., MNLI). 

\section{Conclusion}
We investigated the effects of neural TS systems on downstream NLP tasks using two strategies: (a) simplifying input texts at prediction time, and (b) augmenting data to provide machines with additional information during training. Our experiments indicate that the latter strategy consistently helps multiple NLP tasks, regardless of the underlying method used to address the task, or the neural approach used for TS. 
\section*{Acknowledgements}
This work was supported by the Defense Advanced Research Projects Agency (DARPA) under the World Modelers and HABITUS programs. Mihai Surdeanu declares a financial interest in lum.ai. This interest has been properly disclosed to the University of Arizona Institutional Review Committee, and is managed in accordance with its conflict of interest policies.

\bibliography{anthology,custom}
\bibliographystyle{acl_natbib}




\end{document}